\documentclass{article}

 \usepackage[preprint]{neurips_2026}

\usepackage{marvosym}

\usepackage[utf8]{inputenc} 
\usepackage[T1]{fontenc}    
\usepackage{hyperref}       
\usepackage{url}            
\usepackage{booktabs}       
\usepackage{amsfonts}       
\usepackage{nicefrac}       
\usepackage{microtype}      
\usepackage{xcolor}         
\usepackage{color}
\usepackage{multirow,multicol,booktabs}
\usepackage{multicol}
\usepackage{caption}
\usepackage{adjustbox}
\usepackage{makecell}
\usepackage{ulem}
\usepackage[table]{xcolor}
\usepackage{tabularx}
\usepackage{enumitem}
\usepackage[most]{tcolorbox}
\tcbuselibrary{breakable}
\usepackage{natbib}
\setcitestyle{numbers,square,citesep={,}}

\usepackage{subcaption}

\hypersetup{
    colorlinks=true,
    citecolor=cyan, 
    linkcolor=red,  
    urlcolor=cyan,
}

\definecolor{Gray}{gray}{0.95}

\newcommand{\bs}[1]{{\textbf{#1}}}

\newcommand{\method}{MemCoRe}

\title{MemCoRe: Recovering Evidence from Progressively Compressed Factual Knowledge for Agent Memory}

\author{%
  Zhenyuan Zhang$^{1}$, Xianzhang Jia$^{1}$, Zhiqin Yang$^{1}$\\
  \textbf{Zhenbo Song$^{2}$, Wei Xue$^{1}$, Sirui Han$^{1}$, Yike Guo$^{1}$}\thanks{Corresponding Author. Email: yikeguo@ust.hk}\\
  \; \\
  $^{1}$\;The Hong Kong University of Science and Technology\\ $^{2}$\; Nanjing University of Science and Technology
}

\begin{document}

\maketitle

\begin{abstract}
Memory systems enable LLM agents to consolidate and retrieve relevant evidence from the factual knowledge accumulated through growing interaction histories for downstream reasoning. Existing approaches have explored diverse strategies for organizing and compressing these histories. However, balancing compression with retrieval effectiveness remains challenging: retaining too much content can cause relevant evidence to be obscured by redundant entries, while discarding too aggressively may remove content that later proves relevant. This amounts to a tradeoff between compressing redundancy and preserving enough structure to retrieve target evidence, as formalized by the information bottleneck. To this end, we propose \textbf{\textsc{\method}}, which organizes memory as a compression hierarchy where each level compresses redundancy further while retaining the structure needed for retrieval at that level. In this hierarchy, evidence is progressively compressed from detailed records through extracted keywords to topic groups. This enables retrieval to locate target evidence by searching across levels of the hierarchy. Comprehensive experiments demonstrate that \textsc{\method} outperforms existing state-of-the-art baselines. 
\end{abstract}

\section{Introduction}
\label{sec:intro}
Memory systems enable LLM agents to consolidate and retrieve relevant evidence from the factual knowledge accumulated through growing interaction histories for downstream reasoning~\cite{Xi2023TheRA,Wang_2024,ferrag2025llmreasoningautonomousai}. As interaction histories grow, target evidence is progressively outnumbered by accumulated observations, making accurate retrieval increasingly difficult. To manage this growth, agents construct a structured memory representation and retrieve from it rather than searching the full interaction history. The structure and content of this representation directly govern which evidence can be located at retrieval time, making memory construction inseparable from retrieval effectiveness. 

\begin{figure}[t]
    \centering
    \begin{subfigure}[b]{0.41\textwidth}
        \centering
        \includegraphics[width=\textwidth]{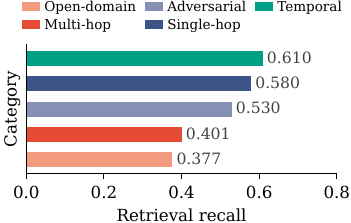}
        \caption{Per-category top-20 retrieval recall. }
        \label{fig:recall}
    \end{subfigure}
    \begin{subfigure}[b]{0.49\textwidth}
        \centering
        \includegraphics[width=\textwidth]{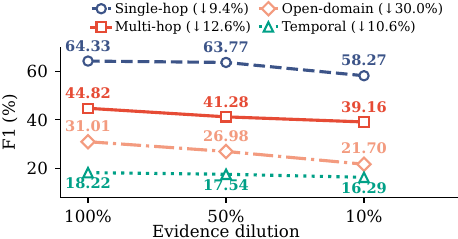}
        \caption{Downstream F1 under evidence dilution. }
        \label{fig:dilution}
    \end{subfigure}
    \vspace{-1mm}
    \caption{Diagnostic analysis on LoCoMo (A-MEM, Qwen3-8B). \textbf{(a)}~Per-category top-20 retrieval recall; 28.2\% of results are near-duplicates. \textbf{(b)}~Downstream F1 when target evidence is diluted to 100\%/50\%/10\% of the retrieved context. Setup details in Appendix~\ref{app:diagnostic}.}
\label{fig:motivation}
\end{figure}

Retrieval-augmented generation has been widely studied for accessing static knowledge bases~\cite{lewisRetrievalAugmentedGenerationKnowledgeIntensive2021,gao2024retrievalaugmentedgenerationlargelanguage}, but agent memory presents a distinct challenge: the retrieval target itself is not given but constructed on the fly from a growing interaction stream. Existing methods have explored diverse strategies for this construction, including structured note generation, associative linking, and selective forgetting~\cite{packerMemGPTLLMsOperating2024,zhongMemoryBankEnhancingLarge2023,xuAMEMAgenticMemory2025,wangOMemOmniMemory2025,zhang2025gmemorytracinghierarchicalmemory,fang2025lightmemlightweightefficientmemoryaugmented}. Our diagnostic analysis on A-MEM~\cite{xuAMEMAgenticMemory2025} (Figure~\ref{fig:recall}) reveals that even when memory retains nearly all raw information, close to half of the ground-truth evidence fails to appear in retrieval results, as redundant entries crowd out relevant content. This retrieval failure varies systematically with query type: open-domain and multi-hop queries, which rely on different granularities of information, suffer the lowest recall. A controlled experiment (Figure~\ref{fig:dilution}) further shows that this gap propagates to downstream reasoning: as target evidence is diluted among irrelevant entries, response quality degrades, with the steepest decline on the same categories that exhibit the lowest retrieval recall. While retaining all content buries evidence, compressing too aggressively risks discarding it altogether. 
This amounts to a tradeoff between compressing redundancy and preserving enough structure to retrieve target evidence, as formalized by the information bottleneck principle~\cite{tishbyInformationBottleneckMethod1999} (Sec.~\ref{sec:problem_formulation}). 

These findings motivate organizing memory as a compression hierarchy. We propose \textsc{\method} (\textbf{Mem}ory \textbf{Co}mpression and \textbf{Re}covery), in which evidence is progressively compressed from detailed records through extracted keywords to topic groups, each retaining the structure needed for retrieval at that level. A gated mechanism consolidates each incoming observation at the evidence level by evaluating its relation to existing content: redundant information is merged, complementary evidence is linked, and novel content is retained. 
This enables retrieval to locate target evidence by searching across levels of the hierarchy: topic groups provide thematic entry points, keywords enable entity-level matching, and associative links connect related evidence for multi-hop reasoning. For queries requiring multiple pieces of evidence, an iterative refinement protocol progressively expands the retrieved set.

The contributions of this work are as follows:
\begin{itemize}[leftmargin=1cm]
\item We conduct a diagnostic analysis revealing that existing memory strategies fail to surface close to half of target evidence at retrieval time, and that this retrieval failure propagates to downstream reasoning with category-dependent severity. The information bottleneck provides a principled framework for understanding these failures as a compression-retrievability tradeoff.

\item We propose \textsc{\method}, which organizes memory as a compression hierarchy where evidence is progressively compressed from detailed records through keywords to topic groups. A gated construction mechanism consolidates observations at the evidence level, and retrieval operates across the hierarchy, with iterative refinement for multi-hop queries.

\item Comprehensive experiments with four backbone models show consistent improvements over state-of-the-art baselines, with the largest gains on the query categories identified as most challenging in the diagnostic analysis. 
\end{itemize}
\begin{figure*}[t] 
    \centering
  \includegraphics[width=0.95\textwidth]{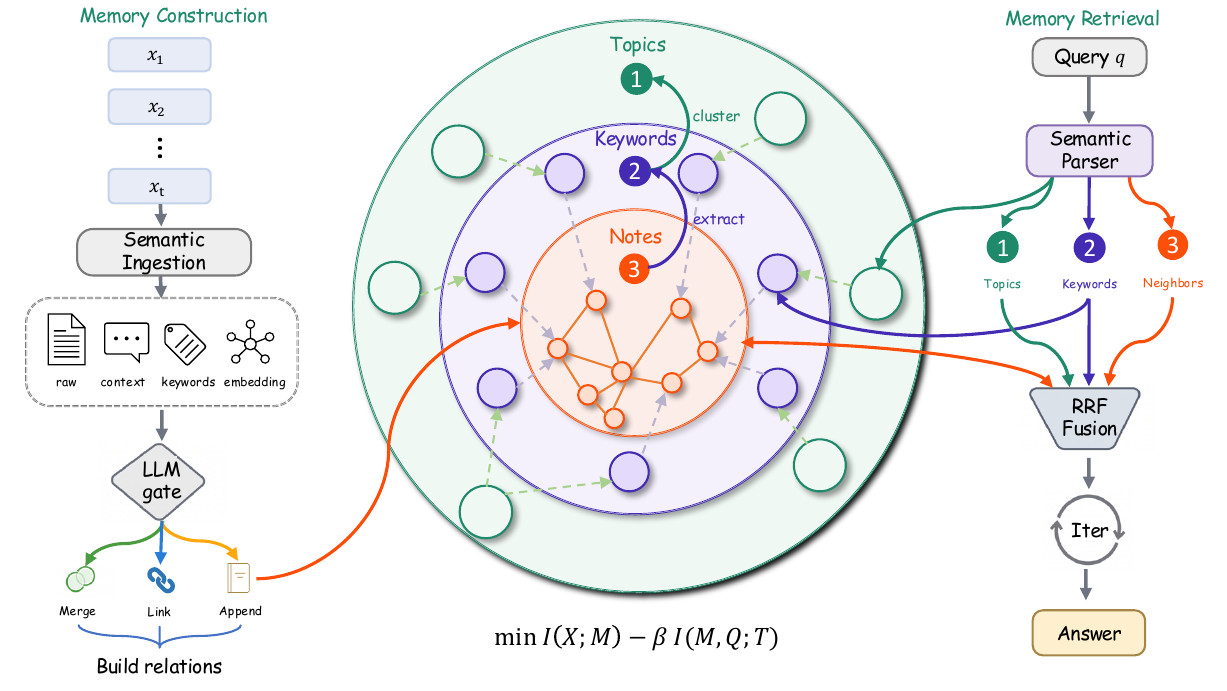}
  \caption{Overview of the \textsc{\method} framework. \textbf{Left}: Memory construction processes incoming observations through semantic ingestion and gated construction, where each observation is merged, linked, or appended based on its semantic relation to existing memory. \textbf{Center}: The memory state is organized as a compression hierarchy spanning notes, keywords, and topic groups, with associative edges connecting related notes. \textbf{Right}: Memory retrieval searches across levels of the hierarchy via topic, keyword, and neighbor expansion channels, followed by iterative evidence refinement for complex queries.} 
  \label{fig:pipeline}
\end{figure*}

\section{The \textsc{\method} Framework}
The diagnostic analysis in Section~\ref{sec:intro} identifies a fundamental tension in memory construction: retaining too much content buries evidence, while compressing too aggressively discards it. The information bottleneck principle~\cite{tishbyInformationBottleneckMethod1999} formalizes this tension as a tradeoff between compressing redundancy and preserving enough structure to retrieve target evidence. Guided by this formalization, \textsc{\method} organizes memory as a compression hierarchy where each level compresses redundancy further while retaining the structure needed for retrieval at that level. Construction builds this hierarchy incrementally (Sec.~\ref{sec:online_consolidation}), and retrieval operates across its levels to locate target evidence (Sec.~\ref{sec:retrieval}). Figure~\ref{fig:pipeline} illustrates the overall architecture.

\subsection{Problem Formulation}
\label{sec:problem_formulation}

We consider the problem of agent memory: locating query-dependent evidence from the factual knowledge accumulated through a growing interaction history. An agent observes a stream of interactions $X_{1:t} = \{x_1, \dots, x_t\}$. When a query $Q$ arrives, the goal is to locate target evidence $T \subseteq X_{1:t}$. To manage retrieval as $|X_{1:t}|$ grows, the agent maintains a memory representation $M_t$ constructed incrementally from $X_{1:t}$. The central challenge is to construct $M_t$ such that it compresses $X_{1:t}$ while preserving the structure needed to recover $T$ for diverse queries.

\paragraph{Adapting the Information Bottleneck. }
The standard information bottleneck framework~\cite{tishbyInformationBottleneckMethod1999} characterizes a compressed representation $Z$ of input $X$ for predicting target $Y$ through the Lagrangian $I(X; Z) - \beta \, I(Z; Y)$. 
We adapt this formulation to the memory retrieval setting with two modifications. The target variable $Y$ is replaced by the pair $(Q, T)$, and the relevance term becomes $I(M_t, Q;\,T)$, reflecting that locating evidence is not a property of memory alone: the same memory state may surface different evidence depending on the query. The adapted Lagrangian is:
\begin{equation}
    \mathcal{L}(M_t) = 
    I(X_{1:t};\, M_t) - \beta \, I(M_t, Q;\, T),
    \label{eq:ib}
\end{equation}
where $\beta > 0$ controls the tradeoff between compression and retrievability. The first term $I(X_{1:t}; M_t)$ measures how much of the raw interaction history is retained in memory: lower values indicate a more compact representation with less redundancy. The second term $I(M_t, Q; T)$ measures how well the memory representation and query jointly support locating target evidence. A favorable memory representation achieves low $\mathcal{L}(M_t)$ by compressing redundancy while preserving retrievability. 
This adapted formulation provides a formal vocabulary for the retrieval failures identified in our diagnostic analysis: memory that retains too much without consolidation exhibits high $I(X; M)$ but low $I(M, Q; T)$, as redundant content interferes with evidence retrieval. 

\paragraph{From Analysis to Design.}
The diagnostic finding that retrieval failures vary systematically across query types motivates organizing memory at multiple levels of compression. In \textsc{\method}, evidence is progressively compressed from detailed records through extracted keywords to topic groups (Section~\ref{sec:online_consolidation}). 
Within this hierarchy, the effect of incorporating a new observation $x_{t+1}$ at the evidence level can be analyzed through the change in $\mathcal{L}$:
\begin{equation} 
    \Delta \mathcal{L} = 
    \underbrace{I(X_{1:t+1}; M_{t+1}) 
    - I(X_{1:t}; M_t)}_{\Delta I_{\text{compress}}} 
    \;-\; \beta
    \underbrace{\bigl(I(M_{t+1}, Q; T) 
    - I(M_t, Q; T)\bigr)}_{\Delta I_{\text{retrieve}}}.
\label{eq:incremental}
\end{equation} 

When $x_{t+1}$ largely duplicates an existing entry, \textbf{merging} them does not introduce new content ($\Delta I_{\text{compress}} \leq 0$), and since the existing entry already covers $x_{t+1}$'s contribution to retrieval, $\Delta I_{\text{retrieve}} \approx 0$. When $x_{t+1}$ is related to an existing entry but carries distinct information, \textbf{linking} them preserves both entries while encoding their relationship, contributing positively to $\Delta I_{\text{retrieve}}$ for queries requiring multi-hop reasoning. When $x_{t+1}$ is independent of all existing entries, \textbf{appending} it increases $\Delta I_{\text{compress}}$ but is necessary to maintain coverage, as $x_{t+1}$ may be the sole evidence for future queries. 
 
Computing mutual information in text space is intractable. However, the design principles above depend on assessing the semantic relation between a new observation and existing memory, specifically whether the new content is redundant, complementary, or novel. Large language models, through pre-trained semantic understanding, provide a natural mechanism for this assessment (Section~\ref{sec:gated_update}). Ablation results (Section~\ref{sec:ablation}) validate the contribution of each operation to downstream performance.

\subsection{Memory Construction}
\label{sec:online_consolidation}
The compression hierarchy in \textsc{\method} consists of three levels (Figure~\ref{fig:pipeline}): notes store detailed evidence and denoised context at the finest level, keywords capture entity-level information as an intermediate compression, and topic groups cluster related keywords at the coarsest level. Associative links record semantic relationships between notes. Construction builds this hierarchy incrementally through three stages: semantic ingestion transforms each raw observation into a structured note, gated construction consolidates the note into existing memory, and topic evolution maintains topic groups over the keyword layer.

\paragraph{Semantic Ingestion.}
\label{sec:ingestion}
Raw observations often contain elliptical references, syntactic noise, and implicit context. We transform each raw observation $x_t$ into a structured note $n_t$ via an LLM-based ingestion function:
\begin{equation}
    n_t = \mathcal{F}_{\text{ingest}}(x_t) = (r_t, c_t, \mathbf{h}_t, \mathcal{K}_t),
\end{equation}
where $r_t$ preserves the verbatim content, $c_t$ is a denoised context summary generated to enhance retrieval relevance, $\mathbf{h}_t = \text{Embed}(c_t) \in \mathbb{R}^d$ is a dense embedding of the context, and $\mathcal{K}_t \subset \mathcal{K}$ is a set of keywords extracted from $x_t$. Retaining both $r_t$ and $c_t$ ensures that raw fidelity is preserved while retrieval operates over cleaner representations. Keywords provide an intermediate symbolic layer between notes and topic groups, grounding proximity in shared symbolic substructures rather than relying solely on embedding similarity.

\paragraph{Gated Construction.}
\label{sec:gated_update}
Following the design principles in Section~\ref{sec:problem_formulation}, which formalize the diagnostic finding that construction decisions depend on the semantic relation between new content and existing memory, each new note $n_t$ is consolidated through one of three operations. The process begins by retrieving a candidate set $\mathcal{N}_{\text{cand}}$ from existing memory through embedding similarity, restricting the evaluation to the most relevant entries. 
An LLM then evaluates each candidate pair $(n_t, n_i)$ with $n_i \in \mathcal{N}_{\text{cand}}$, classifying their semantic relation and estimating a connection strength $s(n_t, n_i) \in [0, 1]$ through structured prompting. Based on this evaluation, the mechanism selects one of three operations:
\begin{equation}
    M_{t+1} \leftarrow \mathcal{O}(n_t, n_i) = 
    \begin{cases}
        \textsc{Merge}(n_i \leftarrow n_i \oplus n_t) & \text{if redundant} \\
        \textsc{Link}(n_i \leftrightarrow n_t) & \text{if complementary} \\
        \textsc{Append}(n_t) & \text{if novel}
    \end{cases}.
\label{eq:update_rule}
\end{equation}
 
\noindent\textit{Merge.} When $n_t$ largely duplicates an existing entry $n_i$, their content is consolidated into a single note:
\begin{equation}
    n_i \leftarrow \bigl(r_i \cup r_t,\;\; 
    \mathcal{F}_{\text{merge}}(c_i, c_t),\;\; 
    \text{Embed}(c'_i),\;\; 
    \mathcal{K}_i \cup \mathcal{K}_t\bigr),
\label{eq:merge}
\end{equation}
where $\mathcal{F}_{\text{merge}}$ is an LLM-based function that synthesizes a unified context $c'_i$ preserving all distinct information from both notes. Raw content from both notes is retained to maintain fidelity, while the merged context reduces redundancy in the retrieval-facing representation.

\noindent\textit{Link.} When $n_t$ is related to $n_i$ but carries distinct information, a new note is created for $n_t$ and an associative edge is established between them:
\begin{equation}
    \mathcal{E}_{\textsc{Related}} \leftarrow \mathcal{E}_{\textsc{Related}} \cup \{(n_t, n_i, \; \rho, \; s)\},
\end{equation}
where $\rho \in \{\textsc{Supports}, \textsc{Conflicts}, \textsc{Related\_To}\}$ denotes the relation type and $s \in [0, 1]$ is the connection strength, both estimated by the LLM. For high-strength connections, the context of the neighboring note $n_i$ is updated to incorporate relevant information from $n_t$, enabling partial information transfer without full merging. 
 
\noindent\textit{Append.} When $n_t$ is independent of all existing entries, it is added as a new autonomous note, preserving coverage for potentially novel evidence. 
 
\paragraph{Topic Evolution.}
\label{sec:topic_evolution}
Keywords extracted from the same note are connected by co-occurrence edges, forming a keyword graph $\mathcal{G}_{\text{kw}}$. As new notes and keywords accumulate, the topic layer is periodically restructured by partitioning this graph:
\begin{equation}
    \max_{\mathcal{T}} \; \mathcal{Q}(\mathcal{T}, \mathcal{G}_{\text{kw}})
    \quad \text{s.t.} \quad \delta_{\min} \leq |C_i| \leq \delta_{\max}, \; \forall C_i \in \mathcal{T},
\end{equation}
where $\mathcal{T} = \{C_1, \dots, C_T\}$ is the set of topic clusters, $\mathcal{Q}$ denotes the modularity function, and $\delta_{\min}, \delta_{\max}$ are cardinality bounds. We employ the Leiden algorithm~\cite{leiden2019} for computational efficiency. Each topic $C_i$ serves as a centroid embedding over its constituent keywords.

Together, these three stages produce a memory state equipped with keywords, topic groups, and weighted associative links, each serving as input to a dedicated retrieval channel (Section~\ref{sec:retrieval}).

\subsection{Memory Retrieval}
\label{sec:retrieval}
The compression hierarchy produced during construction (Section~\ref{sec:online_consolidation}) directly enables multi-level retrieval. Topic groups provide entry points for thematic queries, keywords enable entity-level matching, and associative links connect related evidence for multi-hop reasoning. For queries requiring evidence not directly accessible from initial retrieval, an iterative refinement protocol progressively expands the evidence pool.

\paragraph{Query Decomposition.}
Given a raw query $q$, an LLM-based semantic parser extracts two complementary signals: a topical description embedding $\mathbf{h}_{\text{topic}} \in \mathbb{R}^d$ that captures the query's thematic intent, and a set of entity embeddings $\mathbf{H}_{\text{keys}} = \{\mathbf{h}_{k_1}, \ldots, \mathbf{h}_{k_m}\}$ corresponding to the core entities mentioned in the query. These two signals drive the topic and keyword channels respectively.  

\paragraph{Tri-Pathway Retrieval.}
The \textit{topic channel} operates at the coarsest level of the hierarchy. Given $\mathbf{h}_{\text{topic}}$, we identify the top-$K_{\text{topic}}$ topic groups whose centroid embeddings are most similar to the query's thematic description, then retrieve all notes whose keywords belong to those groups. This channel provides broad thematic coverage, particularly for queries that do not reference specific entities.

The \textit{keyword channel} operates at the intermediate level. Each keyword $k_j \in \mathcal{K}$ maintains a dense embedding $\mathbf{e}_j \in \mathbb{R}^d$; the top-$K_{\text{key}}$ keywords matching each entity embedding in $\mathbf{H}_{\text{keys}}$ are identified by cosine similarity, and all notes containing those keywords are retrieved. This channel provides precise entity-level access, particularly for queries that reference specific people, places, or events.
 
The \textit{neighbor expansion} channel operates at the evidence level. Starting from the anchor set $\mathcal{E}_{\text{anc}} = \mathcal{R}_{\text{topic}} \cup \mathcal{R}_{\text{key}}$ produced by the first two channels, we expand along the associative edges $\mathcal{E}_{\textsc{Related}}$ established during gated construction:
\begin{equation}
    \mathcal{E}_{\text{expand}} = \{m \in \mathcal{N} \mid \exists\, n \in \mathcal{E}_{\text{anc}},\; (n, m) \in \mathcal{E}_{\textsc{Related}}\}.
\end{equation}
Connection strength stored on each edge modulates score propagation, prioritizing neighbors with stronger semantic relationships. This channel retrieves evidence that is logically related to anchor notes but may not share surface-level similarity with the query. 
Results from the topic and keyword channels are combined via Reciprocal Rank Fusion (RRF)~\cite{rrf2009}, which aggregates reciprocal ranks across channels without requiring score normalization. The fused ranking is merged with the expanded neighbor set, and the top-$K_{\text{final}}$ notes form the final evidence pool:
\begin{equation}
\label{eq:evidence_pool}
    \mathcal{E}_{\text{pool}} = \operatorname{Top\text{-}}{K_{\text{final}}}\bigl(
    \text{score}_{\text{RRF}} \cup \mathcal{E}_{\text{expand}}\bigr).
\end{equation}

\paragraph{Iterative Evidence Refinement.}
\label{sec:iterative_refinement}
Complex queries may require evidence not directly accessible from the initial retrieval. We address this through an Iterative Evidence Refinement (IER) protocol that progressively expands the evidence pool. At each iteration, an LLM evaluates whether the current evidence pool sufficiently addresses the query. If gaps are identified, a refined sub-query is synthesized to target missing aspects, and retrieval is re-executed via the tri-pathway mechanism. Newly retrieved notes that are not already in the pool are added, and the process repeats until the LLM judges coverage as sufficient or the maximum iteration count $I_{\max}$ is reached. 
\section{Experiments}
\label{sec:exp}

\begin{table*}[tb!]
    \centering
    \caption{Main results on LoCoMo benchmark using closed-source models (GPT series). We report F1 and BLEU-1 (\%) scores across five categories. The best performance in each category is marked in \textbf{bold}, and the \uline{second best} is underlined. }
    \label{tab:main_gpt}
    \resizebox{\textwidth}{!}{%
    \begin{tabular}{cl|cccccccccc|cc}
    \hline
    \multicolumn{1}{c}{\multirow{3}{*}{\textbf{Model}}} & \multicolumn{1}{c|}{\multirow{3}{*}{\textbf{Method}}} & \multicolumn{10}{c|}{\textbf{Category}} & \multicolumn{2}{c}{\textbf{Average}} \\ \cline{3-14} 
    \multicolumn{1}{c}{} & \multicolumn{1}{c|}{} & \multicolumn{2}{c|}{\textbf{Multi Hop}} & \multicolumn{2}{c|}{\textbf{Temporal}} & \multicolumn{2}{c|}{\textbf{Open Domain}} & \multicolumn{2}{c|}{\textbf{Single Hop}} & \multicolumn{2}{c|}{\textbf{Adversarial}}  & \multirow{2}{*}{\textbf{F1}} & \multirow{2}{*}{\textbf{BLEU}}\\
    \multicolumn{1}{c}{} & \multicolumn{1}{c|}{} & \textbf{F1} & \multicolumn{1}{c|}{\textbf{BLEU}} & \textbf{F1} & \multicolumn{1}{c|}{\textbf{BLEU}} & \textbf{F1} & \multicolumn{1}{c|}{\textbf{BLEU}} & \textbf{F1} & \multicolumn{1}{c|}{\textbf{BLEU}} & \textbf{F1} & \multicolumn{1}{c|}{\textbf{BLEU}} &  &  \\ \hline
    
    \multicolumn{1}{c|}{\multirow{8}{*}{\textbf{\rotatebox{90}{GPT-4o-mini}}}} & \textsc{LoCoMo} & 25.02 & \multicolumn{1}{c|}{19.75} & 18.41 & \multicolumn{1}{c|}{14.77} & 12.04 & \multicolumn{1}{c|}{11.16} & 40.36 & \multicolumn{1}{c|}{29.05} & \bs{69.23} & \multicolumn{1}{c|}{\bs{68.75}} & 39.74 & 33.47 \\

    \multicolumn{1}{c|}{} & \textsc{ReadAgent} & 9.15 & \multicolumn{1}{c|}{6.48} & 12.60 & \multicolumn{1}{c|}{8.87} & 5.31 & \multicolumn{1}{c|}{5.12} & 9.67 & \multicolumn{1}{c|}{7.66} & 9.81 & \multicolumn{1}{c|}{9.02} & 9.89 & 7.87 \\

    \multicolumn{1}{c|}{} & \textsc{MemoryBank} & 5.00 & \multicolumn{1}{c|}{4.77} & 9.68 & \multicolumn{1}{c|}{6.99} & 5.56 & \multicolumn{1}{c|}{5.94} & 6.61 & \multicolumn{1}{c|}{5.16} & 7.36 & \multicolumn{1}{c|}{6.48} & 6.99 & 5.73 \\
     
    \multicolumn{1}{c|}{} & \textsc{MemGPT} & 26.65 & \multicolumn{1}{c|}{17.72} & 25.52 & \multicolumn{1}{c|}{19.44} & 9.15 & \multicolumn{1}{c|}{7.44} & 41.04 & \multicolumn{1}{c|}{34.34} & 43.29 & \multicolumn{1}{c|}{42.73} & 35.45 & 30.16 \\
     
    \multicolumn{1}{c|}{} & \textsc{A-mem} & 27.02 & \multicolumn{1}{c|}{20.09} & 45.85 & \multicolumn{1}{c|}{\textbf{36.67}} & 12.14 & \multicolumn{1}{c|}{12.00} & {44.65} & \multicolumn{1}{c|}{37.06} &  50.03 & \multicolumn{1}{c|}{ 49.47} & \uline{41.97} & \uline{36.16}  \\ 

    \multicolumn{1}{c|}{} & \textsc{Mem-0} & \textbf{34.72} & \multicolumn{1}{c|}{\textbf{25.13}} & \uline{45.93} & \multicolumn{1}{c|}{\uline{35.51}} & {22.64} & \multicolumn{1}{c|}{{15.58}} & 43.65 & \multicolumn{1}{c|}{\uline{37.42}} &  30.15 & \multicolumn{1}{c|}{27.44} & 38.70 & 32.07  \\

    \multicolumn{1}{c|}{} & \textsc{LightMem} & 32.09& \multicolumn{1}{c|}{24.14} & 44.84 & \multicolumn{1}{c|}{30.41} & \uline{22.75} & \multicolumn{1}{c|}{\uline{15.89}} & \uline{44.68} & \multicolumn{1}{c|}{36.11} &  30.13 & \multicolumn{1}{c|}{28.69} & 38.59   & 30.85  \\ 

    \multicolumn{1}{c|}{} & \textsc{\method} & \uline{32.11} & \multicolumn{1}{c|}{\uline{24.48}} & \textbf{46.61} & \multicolumn{1}{c|}{{31.84}} & \textbf{23.98} & \multicolumn{1}{c|}{\textbf{16.84}} & \textbf{44.74} & \multicolumn{1}{c|}{\textbf{38.17}} &  {\uline{51.48}} & \multicolumn{1}{c|}{\uline{51.96}} & \textbf{43.76}   & \textbf{37.27}  \\ 
    \hline
    \multicolumn{1}{c|}{\multirow{8}{*}{\textbf{\rotatebox{90}{GPT-4o}}}} & \textsc{LoCoMo} & 28.00 & \multicolumn{1}{c|}{18.47} & 9.09 & \multicolumn{1}{c|}{5.78} & 16.47 & \multicolumn{1}{c|}{14.80} & \textbf{61.56} & \multicolumn{1}{c|}{\textbf{54.19}} & \textbf{52.61} & \multicolumn{1}{c|}{\textbf{51.13}} & \uline{44.12} & \textbf{38.70} \\
    
    \multicolumn{1}{c|}{} & \textsc{ReadAgent} & 14.61 & \multicolumn{1}{c|}{9.95} & 4.16 & \multicolumn{1}{c|}{3.19} & 8.84 & \multicolumn{1}{c|}{8.37} & 12.46 & \multicolumn{1}{c|}{10.29} & 6.81 & \multicolumn{1}{c|}{6.13} & 9.98 & 8.07  \\
     	
    \multicolumn{1}{c|}{} & \textsc{MemoryBank} & 6.49 & \multicolumn{1}{c|}{4.69} & 2.47 & \multicolumn{1}{c|}{2.43} & 6.43 & \multicolumn{1}{c|}{5.30} & 8.28 & \multicolumn{1}{c|}{7.10} & 4.42 & \multicolumn{1}{c|}{3.67} & 6.13 & 5.15  \\
     
    \multicolumn{1}{c|}{} & \textsc{MemGPT} & 30.36 & \multicolumn{1}{c|}{22.83} & 17.29 & \multicolumn{1}{c|}{13.18} & 12.24 & \multicolumn{1}{c|}{11.87} & \uline{60.16} & \multicolumn{1}{c|}{\uline{53.35}} & 34.96 & \multicolumn{1}{c|}{34.25} & 41.02& \uline{36.23}  \\
     	 
    \multicolumn{1}{c|}{} & \textsc{A-mem} & 32.86 & \multicolumn{1}{c|}{23.76} & 39.41 & \multicolumn{1}{c|}{\uline{31.23}} & 17.10 & \multicolumn{1}{c|}{15.84} &  48.43 & \multicolumn{1}{c|}{ 42.97} &  36.35 & \multicolumn{1}{c|}{ 35.53} & 40.53 & 35.36  \\
     	 
    \multicolumn{1}{c|}{} & \textsc{Mem-0} & {35.13} & \multicolumn{1}{c|}{{27.56}} & \textbf{52.38} & \multicolumn{1}{c|}{\textbf{44.15}} & {17.73} & \multicolumn{1}{c|}{{15.92}} & 39.12 & \multicolumn{1}{c|}{35.43} &  25.44 & \multicolumn{1}{c|}{24.19} & 36.59 & 32.25  \\ 
    										 	 
    \multicolumn{1}{c|}{} & \textsc{LightMem} & \uline{35.21} & \multicolumn{1}{c|}{\uline{30.25}} & \uline{40.28} & \multicolumn{1}{c|}{26.44} & \uline{24.57} & \multicolumn{1}{c|}{\uline{18.32}} & 51.83 & \multicolumn{1}{c|}{39.2} &  29.66 & \multicolumn{1}{c|}{21.52} & 41.31 & 30.89 \\ 
    
\multicolumn{1}{c|}{} & \textsc{\method} & \textbf{35.89} & \multicolumn{1}{c|}{\textbf{29.24}} & {39.78} & \multicolumn{1}{c|}{27.12} & \textbf{25.74} & \multicolumn{1}{c|}{\textbf{19.53}} & 49.08 & \multicolumn{1}{c|}{43.05} & \uline{48.24} & \multicolumn{1}{c|}{\uline{48.92}} & \textbf{44.39} & \textbf{38.70} \\ 
    \hline

    \end{tabular}%
    }
    \end{table*}

\subsection{Experimental Setup}
\label{sec:setup}
 
\noindent\textbf{Dataset.}
We evaluate \textsc{\method} on the LoCoMo benchmark~\cite{maharanaEvaluatingVeryLongTerm2024}, a dataset designed to assess the long-term information synthesis capabilities of LLM agents. LoCoMo contains long-horizon conversations with interleaved topics and evolving entity states, making it a robust testbed for dynamic memory structures. To provide a granular analysis of memory performance, we evaluate on five distinct reasoning categories: Multi-Hop, Temporal, Open Domain, Single Hop, and Adversarial. More details can be found in Appendix~\ref{app:dataset}.

\noindent{\textbf{{Evaluation Metrics.}}}
Following standard evaluation metrics~\cite{xuAMEMAgenticMemory2025}, we employ two primary metrics: F1 Score to measure the token-level overlap and precision of the answer spans, and BLEU-1~\cite{bleu1} to evaluate the lexical fidelity of the generated responses against ground truth. For ablation studies, we additionally report Recall, measuring the proportion of ground-truth evidence retrieved, and Hit Rate, indicating whether any relevant evidence appears in the candidates.

\noindent{\textbf{Implementation Details. }}
We implement \textsc{\method} using a graph-backed architecture with Neo4j, maintaining the compression hierarchy (notes, keywords, and topic groups) as an integrated graph with both vector indices and explicit edges. All embeddings use \texttt{Qwen3-Embedding-0.6B}. For memory construction, the LLM classifies the semantic relation between each new observation and candidate entries, and estimates connection strength $s \in [0, 1]$ through structured prompting (Section~\ref{sec:gated_update}). Topic evolution uses the Leiden algorithm with cluster size bounds $\delta_{\min} = 3$ and $\delta_{\max} = 15$. For retrieval, we set $K_{\text{topic}} = 3$ for topic-based retrieval, $K_{\text{key}} = 10$ for keyword matching, and $K_{\text{final}} = 20$ for the final evidence pool budget. Neighbor expansion performs 1-hop traversal along $\mathcal{E}_{\textsc{Related}}$ edges. The iterative evidence refinement protocol uses $I_{\max} = 3$ iterations. Full hyperparameter settings are provided in Appendix~\ref{app:hyperparameters}.

\noindent\textbf{Backbone Models and Baselines.}
We evaluate \textsc{\method} across four foundation models spanning closed-source (GPT~\cite{openai2024gpt4technicalreport}) and open-source (Qwen~\cite{yang2025qwen3technicalreport}) families: GPT-4o-mini, GPT-4o, Qwen3-8B, and Qwen3-14B. The generation temperature is set to $0.7$ for general reasoning and $0.5$ for adversarial tasks. We compare against seven representative methods: \textsc{LoCoMo}~\cite{maharanaEvaluatingVeryLongTerm2024}, \textsc{ReadAgent}~\cite{leeHumanInspiredReadingAgent2024}, \textsc{MemoryBank}~\cite{zhongMemoryBankEnhancingLarge2023}, \textsc{MemGPT}~\cite{packerMemGPTLLMsOperating2024}, \textsc{A-MEM}~\cite{xuAMEMAgenticMemory2025}, \textsc{Mem0}~\cite{chhikaraMem0BuildingProductionReady2025}, and \textsc{LightMem}~\cite{fang2025lightmemlightweightefficientmemoryaugmented}. All baselines are implemented using their official system prompts and default configurations to ensure a fair comparison.

\subsection{Main Results}
\label{sec:main_results}
\begin{table*}[t]
    \centering
    \caption{
        Main results on LoCoMo benchmark using open-source models (Qwen series). We report F1 and BLEU-1 (\%) scores across five reasoning categories. The best performance in each category is marked in \textbf{bold}, and the \uline{second best} is underlined. 
    }
    \label{tab:main_qwen}
    \resizebox{\textwidth}{!}{%
    \begin{tabular}{cl|cccccccccc|cc}
    \hline
    \multicolumn{1}{c}{\multirow{3}{*}{\textbf{Model}}} & \multicolumn{1}{c|}{\multirow{3}{*}{\textbf{Method}}} & \multicolumn{10}{c|}{\textbf{Category}} & \multicolumn{2}{c}{\textbf{Average}} \\ \cline{3-14} 
    \multicolumn{1}{c}{} & \multicolumn{1}{c|}{} & \multicolumn{2}{c|}{\textbf{Multi Hop}} & \multicolumn{2}{c|}{\textbf{Temporal}} & \multicolumn{2}{c|}{\textbf{Open Domain}} & \multicolumn{2}{c|}{\textbf{Single Hop}} & \multicolumn{2}{c|}{\textbf{Adversarial}}  & \multirow{2}{*}{\textbf{F1}} & \multirow{2}{*}{\textbf{BLEU}}\\
    \multicolumn{1}{c}{} & \multicolumn{1}{c|}{} & \textbf{F1} & \multicolumn{1}{c|}{\textbf{BLEU}} & \textbf{F1} & \multicolumn{1}{c|}{\textbf{BLEU}} & \textbf{F1} & \multicolumn{1}{c|}{\textbf{BLEU}} & \textbf{F1} & \multicolumn{1}{c|}{\textbf{BLEU}} & \textbf{F1} & \multicolumn{1}{c|}{\textbf{BLEU}} &  &  \\ \hline
    
    \multicolumn{1}{c|}{\multirow{8}{*}{\textbf{\rotatebox{90}{Qwen3-8B}}}} & \textsc{LoCoMo} & {25.09} & \multicolumn{1}{c|}{15.73} & 32.82  & \multicolumn{1}{c|}{\uline{27.14}} & 14.47 & \multicolumn{1}{c|}{13.35} & 20.18 & \multicolumn{1}{c|}{18.39} & \textbf{46.77} & \multicolumn{1}{c|}{\uline{40.81}} & 28.62  & 24.22 \\	 	 	 	 	 	 	
    \multicolumn{1}{c|}{} & \textsc{ReadAgent} & 13.17  & \multicolumn{1}{c|}{9.30} & \uline{34.91} & \multicolumn{1}{c|}{27.04} & 8.80 & \multicolumn{1}{c|}{7.45} & 26.44 & \multicolumn{1}{c|}{24.83} & 29.98 & \multicolumn{1}{c|}{28.34} & 25.87 & 22.93 \\
     	 	 	 	 	 	 	 	 	 	
    \multicolumn{1}{c|}{} & \textsc{MemoryBank} & 21.25 & \multicolumn{1}{c|}{14.53} & 30.20 & \multicolumn{1}{c|}{21.11} & 11.33 & \multicolumn{1}{c|}{10.53} & 32.75 & \multicolumn{1}{c|}{26.33} & 30.95 & \multicolumn{1}{c|}{30.13} & 29.27 & 23.90 \\
     	 	 	 	 		 	 	 	 	 	
    \multicolumn{1}{c|}{} & \textsc{MemGPT} & 22.13 & \multicolumn{1}{c|}{13.44} & {31.47} & \multicolumn{1}{c|}{22.16} & 14.51 & \multicolumn{1}{c|}{13.54} & 33.49 & \multicolumn{1}{c|}{\uline{34.12}} & 34.58 & \multicolumn{1}{c|}{31.44} & 30.88 & \uline{27.67} \\

    \multicolumn{1}{c|}{} & \textsc{A-mem} & 24.30 & \multicolumn{1}{c|}{16.90} & 34.50 & \multicolumn{1}{c|}{23.10} & 13.10 & \multicolumn{1}{c|}{12.20} & {38.10} & \multicolumn{1}{c|}{{33.30}} &  31.00 & \multicolumn{1}{c|}{ 30.10} & {32.76} & 27.58  \\ 
     	 	 	 	 	 	 	 	 		 	 
    \multicolumn{1}{c|}{} & \textsc{Mem-0} & 23.04 & \multicolumn{1}{c|}{\uline{19.74}} & 29.65 & \multicolumn{1}{c|}{23.16} & \textbf{20.63} & \multicolumn{1}{c|}{\uline{13.75}} & 30.46 & \multicolumn{1}{c|}{25.62} &  26.02 & \multicolumn{1}{c|}{22.48} & 27.80 & 23.11   \\ 	 	 	 	 	 	 	 	

    \multicolumn{1}{c|}{} & \textsc{LightMem} & \uline{27.19} & \multicolumn{1}{c|}{19.53} & 34.40  & \multicolumn{1}{c|}{20.62 } & 15.21  & \multicolumn{1}{c|}{10.07 } & \uline{40.21} & \multicolumn{1}{c|}{30.19} &  35.22  & \multicolumn{1}{c|}{34.79 } & \uline{35.09}   & 27.19  \\  	 		 	 	 	 	 	 
    
    \multicolumn{1}{c|}{} & \textsc{\method} & \textbf{28.24} & \multicolumn{1}{c|}{\textbf{22.76}} & \textbf{38.39} & \multicolumn{1}{c|}{\textbf{33.64}} & \uline{15.43} & \multicolumn{1}{c|}{\textbf{13.81}} & \textbf{42.09} & \multicolumn{1}{c|}{\textbf{36.57}} &  \uline{43.79} & \multicolumn{1}{c|}{\textbf{43.14}} & \textbf{38.62}   & \textbf{34.51}  \\ 
    \hline
        
    \multicolumn{1}{c|}{\multirow{8}{*}{\textbf{\rotatebox{90}{Qwen3-14B}}}} & \textsc{LoCoMo} & \textbf{33.37} & \multicolumn{1}{c|}{\textbf{24.26}} &\textbf{31.49} & \multicolumn{1}{c|}{16.42} & \uline{13.92} & \multicolumn{1}{c|}{11.02} & 25.46 & \multicolumn{1}{c|}{24.82} & \textbf{49.17} & \multicolumn{1}{c|}{\textbf{35.00}} & \uline{32.42} & 25.02 \\
    										 	 
    \multicolumn{1}{c|}{} & \textsc{ReadAgent} & 13.16 & \multicolumn{1}{c|}{9.61} & 18.12 & \multicolumn{1}{c|}{12.33} & 12.16 & \multicolumn{1}{c|}{9.25} & 32.83 & \multicolumn{1}{c|}{28.35} & 5.96 & \multicolumn{1}{c|}{4.2} & 20.63 & 16.75 \\
        
    \multicolumn{1}{c|}{} & \textsc{MemoryBank} & 25.97 & \multicolumn{1}{c|}{18.16} & 25.37 & \multicolumn{1}{c|}{18.76} & 13.52 & \multicolumn{1}{c|}{11.69} & 34.92 & \multicolumn{1}{c|}{30.6} & 21.94 & \multicolumn{1}{c|}{17.56} & 28.16 & 23.08  \\
    										 	 
    \multicolumn{1}{c|}{} & \textsc{MemGPT} & 24.12 & \multicolumn{1}{c|}{15.41} & 25.48 & \multicolumn{1}{c|}{19.04} & 13.44 & \multicolumn{1}{c|}{\uline{12.64}} & 34.74 & \multicolumn{1}{c|}{\uline{32.41}} & \uline{27.11} & \multicolumn{1}{c|}{24.32} & 28.99 & \uline{25.06}  \\
    										 	 
    \multicolumn{1}{c|}{} & \textsc{A-mem} & 21.36 & \multicolumn{1}{c|}{14.98} & 23.06 & \multicolumn{1}{c|}{18.04} & 12.62 & \multicolumn{1}{c|}{11.49} &  {35.43}& \multicolumn{1}{c|}{ 30.92} &  26.71 & \multicolumn{1}{c|}{\uline{25.78}} & 28.37  &24.48  \\
    											 
    \multicolumn{1}{c|}{} & \textsc{Mem-0} & 20.98 & \multicolumn{1}{c|}{16.27} & 31.5 & \multicolumn{1}{c|}{\uline{21.73}} & 12.7& \multicolumn{1}{c|}{\textbf{13.22}} & 24.7 & \multicolumn{1}{c|}{19.14} &  21.01 & \multicolumn{1}{c|}{19.84} & 23.86 & 19.02  \\ 
    										 	 
    \multicolumn{1}{c|}{} & \textsc{LightMem} & 29.02 & \multicolumn{1}{c|}{22.23} &28.24 & \multicolumn{1}{c|}{16.37} & 13.88 & \multicolumn{1}{c|}{12.32} & \uline{38.41} & \multicolumn{1}{c|}{28.46} &  26.26 & \multicolumn{1}{c|}{24.67} & 31.52 & 23.99 \\ 
    										 	 
    \multicolumn{1}{c|}{} & \textsc{\method} & \uline{30.80} & \multicolumn{1}{c|}{\uline{23.13}} & \uline{29.25} & \multicolumn{1}{c|}{\textbf{24.56}} & \textbf{14.11} & \multicolumn{1}{c|}{11.03} & \textbf{42.25} & \multicolumn{1}{c|}{\textbf{35.52}} & 26.59 & \multicolumn{1}{c|}{25.02} & \textbf{33.65}   & \textbf{28.45}  \\ 
    \hline
    \end{tabular}%
    }
    \end{table*}
\noindent\textbf{Overall Performance.}
Tables~\ref{tab:main_gpt} and~\ref{tab:main_qwen} present performance comparisons on closed-source and open-source models, respectively. \textsc{\method} achieves the highest average F1 across all four backbone models. On GPT-4o-mini, it attains 43.76\% F1, outperforming the second-best \textsc{A-MEM} by 1.79 points. On GPT-4o, \textsc{\method} achieves 44.39\% F1, a narrower margin of 0.27 points over \textsc{LoCoMo}. This gap is small enough that it may not be statistically significant without repeated runs. However,  \textsc{\method} maintains this edge while achieving a substantially more balanced category profile than \textsc{LoCoMo}, whose 52.61\% Adversarial score coexists with only 9.09\% on Temporal. On open-source models, the improvement is particularly notable on Qwen3-8B, where \textsc{\method} reaches 38.62\% F1, surpassing the second-best \textsc{LightMem} by 3.53 points. On Qwen3-14B, \textsc{\method} achieves 33.65\% F1, outperforming all memory-based baselines. The consistent improvements across heterogeneous architectures validate the generalization of our approach. 

\noindent\textbf{Category-wise Analysis.}
\textsc{\method} shows differentiated strengths across query categories. For Single Hop tasks requiring precise entity matching, \textsc{\method} achieves top performance on three of four models, including 42.09\% and 42.25\% F1 on Qwen3-8B and Qwen3-14B respectively. This aligns with the ablation finding that keyword-based retrieval is critical for entity-level queries (Section~\ref{sec:ablation}). For Open Domain queries, \textsc{\method} achieves the highest F1 on both GPT models (23.98\% and 25.74\%), where the topic channel helps localize relevant memory regions. On Multi-Hop queries, \textsc{\method} achieves the best F1 on GPT-4o (35.89\%) and Qwen3-8B (28.24\%), where associative links established during construction provide retrieval paths across related evidence. 


\noindent\textbf{Balance and the Compression Tradeoff.}
Beyond per-category comparisons, \textsc{\method} achieves a more balanced performance profile across categories than baselines that excel in one area but fail in others. For instance, on GPT-4o, \textsc{LoCoMo} reaches 52.61\% on Adversarial but only 9.09\% on Temporal, while \textsc{Mem0} achieves 52.38\% on Temporal but only 25.44\% on Adversarial. In contrast, \textsc{\method} maintains competitive scores across all categories without severe degradation on any. This balance reflects the compression hierarchy design: by providing retrieval structure at multiple levels, the system avoids the extremes of over-retention and over-compression. On GPT models, the largest gains appear on the categories identified as most challenging in the diagnostic analysis. On open-source models, improvements concentrate on Single-Hop and Multi-Hop categories. 
The main exception is Adversarial on Qwen3-14B, where \textsc{LoCoMo} (49.17\%) substantially outperforms \textsc{\method} (26.59\%), indicating that lower-compressed context aids distractor rejection for larger open-source backbones. However, \textsc{\method} achieves the highest average F1 on this backbone. 

\subsection{Ablation Study}
\label{sec:ablation}
\begin{figure}[t]
    \centering
    \begin{minipage}[t]{0.38\textwidth}
        \centering
        \captionof{table}{Ablation study on LoCoMo (Qwen3-8B). 
        The best performance in each category is marked in \textbf{bold}, and the \uline{second best} within each group is underlined.
        }
        \label{tab:ablation}
        \small
        \resizebox{\textwidth}{!}{
        \begin{tabular}{ll|cc|cc}
        \toprule
        \textbf{Phase} & \textbf{Method} & \textbf{F1} & \textbf{BLEU} & \textbf{Recall} & \textbf{Hit Rate}  \\
        \midrule
        \multicolumn{1}{c|}{-} &
        \textsc{\method} & \textbf{38.62} &\textbf{36.85} &\textbf{62.22}&	\textbf{67.12} \\
        \midrule
        \multicolumn{1}{c|}{\multirow{4}{*}{\textbf{\rotatebox{90}{\makecell[c]{Constru-\\ction}}}}}&
        w/o Update & 27.97 & 27.10 & 42.11 & 48.20  \\
        \multicolumn{1}{c|}{}& w/o Denoise& \uline{36.07} & \uline{34.68} & \uline{57.42} & \uline{62.55}  \\
        \multicolumn{1}{c|}{}& w/o Link & 33.57  & 32.35  & 53.19 & 56.18  \\
        \multicolumn{1}{c|}{}& w/o Merge & 34.79  & 33.62  & 54.85 & 59.42  \\
        \midrule
        \multicolumn{1}{c|}{\multirow{4}{*}{\textbf{\rotatebox{90}{Retrieval}}}}&
        w/o Topic & \uline{36.79} & \uline{34.66} & \uline{53.30} & \uline{58.91} \\
        \multicolumn{1}{c|}{}& w/o Keyword & 32.69 & 33.94 & 51.28 & 54.26 \\
        \multicolumn{1}{c|}{}& w/o Neighbor & 34.26 & 32.85 & 51.28 & 54.35  \\
        \multicolumn{1}{c|}{}& w/o IER & 32.94 & 30.86 & 46.29 & 51.26 \\
        \bottomrule
        \end{tabular}
        }
    \end{minipage}
    \hfill
    \begin{minipage}[t]{0.58\textwidth}
        \centering
        \vspace{0pt}
        \includegraphics[width=\textwidth]{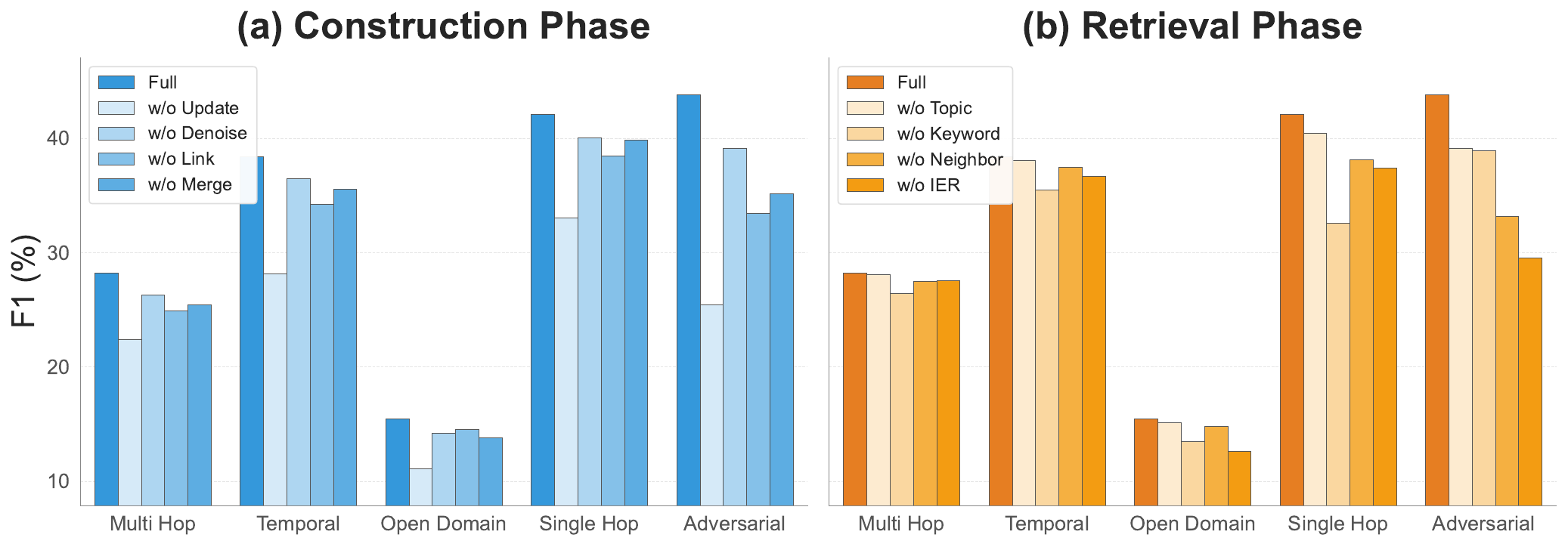}
        \caption{Category-wise F1 scores (\%) for ablation variants on LoCoMo (Qwen3-8B). (a) Ablations on memory construction components. (b) Ablations on retrieval pathways and iterative refinement.}
        \label{fig:ablation_study}
    \end{minipage}
    \vspace{-0.5cm}
\end{figure}

We validate the compression hierarchy by ablating its construction and retrieval components separately. Results are shown in Table~\ref{tab:ablation} and Figure~\ref{fig:ablation_study}. Overall, \textsc{\method} achieves 62.22\% retrieval recall, compared to the 53.8\% observed for \textsc{A-MEM} (the sample-weighted average of the per-category recalls in Figure~\ref{fig:recall}) in the diagnostic analysis (Section~\ref{sec:intro}), recovering a substantially larger fraction of target evidence.

\noindent\textbf{Construction Ablations.}
We first assess how each construction stage contributes to the quality of the compression hierarchy. Removing the entire gated construction (\textit{w/o Update}), which reduces the system to append-only storage without merge or link operations, causes the most severe degradation: average F1 drops from 38.62\% to 27.97\% and Recall from 62.22\% to 42.11\%. As shown in Figure~\ref{fig:ablation_study}(a), this variant exhibits the largest performance gap across all five categories. Without active consolidation, redundant entries accumulate unchecked in the evidence level, directly confirming the compression side of the IB tradeoff: high $I(X; M)$ without structural organization leads to low $I(M, Q; T)$.

Among individual operations, \textit{w/o Link} shows larger impact than \textit{w/o Merge} (33.57\% vs 34.79\% F1). Figure~\ref{fig:ablation_study}(a) reveals that Link removal particularly affects Adversarial performance, indicating that associative edges help preserve relational structure that distinguishes relevant evidence from distractors. The larger impact of Link over Merge suggests that preserving retrievability through relational structure (the $I(M, Q; T)$ term) contributes more than reducing redundancy (the $I(X; M)$ term) at the note level. The \textit{w/o Denoise} variant shows the mildest degradation among construction ablations (36.07\% F1), with relatively stable results across all categories as shown in the figure, indicating that semantic preprocessing provides consistent but auxiliary improvements.

\noindent\textbf{Retrieval Ablations.}
We then assess how each retrieval channel exploits the compression hierarchy. The \textit{w/o Topic} variant achieves the mildest degradation (36.79\% F1), with Figure~\ref{fig:ablation_study}(b) showing relatively uniform impact across categories. This indicates that topic-level retrieval provides general guidance that other channels can partially compensate for. In contrast, \textit{w/o Keyword} (32.69\% F1) and \textit{w/o Neighbor} (34.26\% F1) show more pronounced impacts. As illustrated in Figure~\ref{fig:ablation_study}(b), Keyword removal causes the most pronounced decline on Single-Hop queries, validating that entity-level matching at the intermediate compression level is essential for precise retrieval. The \textit{w/o Neighbor} variant primarily impacts Adversarial queries, confirming that traversal along associative edges helps retrieve evidence that is logically related but not directly similar to the query.

The \textit{w/o IER} variant (32.94\% F1) produces the largest Recall drop among retrieval ablations (62.22\% $\to$ 46.29\%), demonstrating that iterative refinement is critical for reaching evidence not directly accessible from the initial retrieval. Figure~\ref{fig:ablation_study}(b) shows the steepest decline on Adversarial and Open Domain categories, where queries often require progressive evidence accumulation. Multi-hop queries are already partially served by the neighbor expansion channel, which traverses associative links between related evidence; Adversarial and Open Domain queries, which require broader evidence coverage without clear link paths, benefit more from iterative re-retrieval. 

The most severe degradation overall comes from removing gated construction entirely ($-$10.65 F1), indicating that the quality of the compression hierarchy built during construction constrains what retrieval can recover. Notably, link removal (construction) and neighbor expansion removal (retrieval) both primarily impact Adversarial, confirming that the associative structure built during gated construction is what the neighbor channel exploits. Keyword-based retrieval primarily impacts Single-Hop, validating that the entity-level structure produced during ingestion enables precise matching.

\section{Related Work}
\subsection{Memory Systems for LLM Agents}
A growing body of work addresses long-term memory for language agents, with methods differing primarily in how they construct and manage stored representations.

\noindent \textbf{Tiered storage and context management.} MemGPT~\cite{packerMemGPTLLMsOperating2024} introduces an OS-inspired virtual memory hierarchy that pages information between working context and archival storage, enabling effectively unbounded conversation history. HiAgent~\cite{huHiAgentHierarchicalWorking2024} extends this idea with a hierarchical working memory that organizes context at multiple granularities. LightMem~\cite{fang2025lightmemlightweightefficientmemoryaugmented} proposes a lightweight tiered architecture that compresses memory across layers for efficiency. These systems focus on where information resides but leave the decision of what to retain largely to recency or capacity heuristics. 

\noindent \textbf{Forgetting and compression.} MemoryBank~\cite{zhongMemoryBankEnhancingLarge2023} applies the Ebbinghaus forgetting curve to selectively discard older entries, prioritizing recency over completeness. SimpleMem~\cite{simplemem2025} introduces a three-stage pipeline that distills interactions into compact, semantically indexed units. These methods reduce memory footprint but risk discarding content that may later prove relevant.

\noindent \textbf{Associative organization.} A-MEM~\cite{xuAMEMAgenticMemory2025} generates structured notes with keywords and tags for each observation, and establishes associative links inspired by Zettelkasten-style knowledge management. O-Mem~\cite{wangOMemOmniMemory2025} extends this direction with user-centric profiling, while GMemory~\cite{zhang2025gmemorytracinghierarchicalmemory} maintains hierarchical representations that trace memory evolution across sessions. These approaches enrich memory structure but create a new entry for every observation without consolidating redundant content.

\noindent \textbf{Fact extraction.} Mem0~\cite{chhikaraMem0BuildingProductionReady2025} extracts structured facts from interactions and maintains them through add, update, and delete operations over a hybrid vector-graph store. Zep~\cite{zep2025} tracks factual evolution through a temporal knowledge graph. These systems operate at the fact level rather than the evidence level, which may lose contextual detail needed for complex reasoning.
Our diagnostic analysis (Figure~\ref{fig:motivation}) reveals that despite these diverse strategies, close to half of target evidence remains unreachable during retrieval, with the severity varying systematically across query types. 

\subsection{Retrieval-Augmented Generation}
Retrieval-augmented generation (RAG)~\cite{lewisRetrievalAugmentedGenerationKnowledgeIntensive2021,gao2024retrievalaugmentedgenerationlargelanguage} retrieves relevant passages from static corpora to ground language model outputs. Structural approaches organize documents hierarchically through recursive summarization~\cite{sarthiRAPTORRecursiveAbstractive2024} or graph representations~\cite{edgeLocalGlobalGraph2025,wuThinkonGraph30Efficient2025} to support multi-granularity retrieval. Iterative methods refine queries and filter outputs at inference time~\cite{asaiSelfRAGLearningRetrieve2023,yanCorrectiveRetrievalAugmented2024}. These methods optimize retrieval over static corpora, whereas agent memory requires constructing the retrieval target itself from a growing interaction stream, calling for memory representations that jointly support compression and retrieval.
\section{Conclusion}
We presented \textsc{\method}, a memory system for LLM agents built around hierarchical compression. A diagnostic analysis reveals that close to half of target evidence remains unreachable during retrieval, with failures varying systematically across query types. As formalized by the information bottleneck, this amounts to a tradeoff between compression and retrievability. We proposed \textsc{\method}, which organizes memory as a compression hierarchy where evidence is progressively compressed through keywords to topic groups, each level retaining the structure needed for retrieval. A gated construction mechanism populates this hierarchy by consolidating each observation based on its relation to existing content, and retrieval operates across levels of the hierarchy to locate target evidence. Experiments demonstrate consistent improvements over state-of-the-art baselines across four backbone models.

\noindent\textbf{Limitations.}
The current implementation prioritizes memory quality over construction speed, introducing computational overhead from LLM-based semantic evaluation at each construction step. The information bottleneck serves as an analytical framework in this work, however, directly optimizing the IB objective for memory construction is a direction we plan to explore. Extending evaluation to additional benchmarks, multi-modal interactions, and domain-specific scenarios also remains future work.

\normalem
\bibliographystyle{plain}
\bibliography{ref}

\clearpage

\appendix
\onecolumn

\section{Diagnostic Experiment Details}
\label{app:diagnostic}

This section describes the experimental setup for the diagnostic analysis reported in Figure~\ref{fig:motivation}. Both experiments use the LoCoMo benchmark (Appendix~\ref{app:dataset}) and A-MEM~\cite{xuAMEMAgenticMemory2025} as the representative structured memory system.

\paragraph{Shared setup.}
A-MEM processes each conversational turn into a structured note with metadata and associative links, following the Zettelkasten knowledge management method. Running A-MEM on the full LoCoMo dataset produces $5,882$ memory notes across $10$ conversations. All embeddings use \texttt{Qwen3-Embedding-0.6B} for both A-MEM and \textsc{\method} to ensure a fair comparison. 
To identify ground-truth evidence for each query, we parse the dialogue IDs annotated in LoCoMo's evidence field, reconstruct the corresponding utterances, and match them against stored notes by exact content matching. 

\subsection{Experiment 1: Retrieval Diagnosis}
\label{app:exp1}

This experiment characterizes retrieval quality without any LLM inference. For each of the $1,986$ queries, we retrieve the top-20 most similar notes using cosine similarity and measure three quantities:

\begin{enumerate}
    \item \textbf{Storage-level redundancy}: the distribution of pairwise cosine similarity across all 5,882 notes, evaluated at thresholds $0.7$, $0.8$, and $0.9$.
    \item \textbf{Retrieval-level redundancy}: the fraction of note pairs within each top-20 result set whose cosine similarity exceeds $0.8$.
    \item \textbf{Top-20 recall}: the fraction of ground-truth evidence notes that appear in the top-20 results, reported per query category.
\end{enumerate}

Results are computed using note embeddings to isolate the effect of semantic content on retrieval quality.

\subsection{Experiment 2: Evidence Dilution}
\label{app:exp2}

This experiment measures how downstream reasoning quality degrades when target evidence is diluted among irrelevant entries, isolating the causal effect of evidence concentration on task performance.

\paragraph{Dilution protocol.} For each query, we identify its ground-truth notes and control the fraction of target evidence in the input context. Given a target evidence ratio $r \in \{1.0, 0.5, 0.1\}$, we inject $\lfloor |GT| \cdot (1/r - 1) \rceil$ distractor notes, where $|GT|$ is the number of ground-truth notes. Distractors are sampled from non-ground-truth notes within the same conversation and shuffled with the ground-truth notes before concatenation.

\paragraph{Prompt and evaluation.} A category-agnostic prompt (``Based on the following memory notes, answer the question. Be concise.'') is used to isolate dilution as the sole experimental variable. Responses are generated by Qwen3-8B and evaluated by token-set F1 against the gold answer.

\paragraph{Scope.} We evaluate 1,531 queries from categories 1--4 (single-hop, multi-hop, temporal, open-domain). Category~5 (adversarial) is excluded because its gold answers are refusal phrases (e.g., ``Not mentioned in the conversation''). Under the category-agnostic prompt, the model consistently generates substantive answers rather than appropriate refusals, producing near-zero F1 regardless of dilution ratio (F1 $\leq$ 1.04 across all settings). Including this category would conflate prompt design effects with the dilution signal under study.

\section{Detailed Analysis of Construction Operations}
\label{app:ib_derivation}

This appendix provides a more detailed analysis of how each construction operation affects the two quantities in the IB Lagrangian (Eq.~\eqref{eq:ib}), expanding on the design principles presented in Section~\ref{sec:problem_formulation}.
 
\subsection{Incremental Analysis}
 
Let $m^* = \arg\max_{m_j \in M_t} \text{sim}(x_t, m_j)$ denote the existing memory entry most similar to the new observation $x_t$. We analyze each operation through its effect on the compression term $I(X_{1:t}; M_t)$ and the retrievability term $I(M_t, Q; T)$.
 
\paragraph{Merge.}
When $x_t$ is highly redundant with $m^*$, merging them into a single entry $\tilde{m}$ replaces $m^*$ without increasing the number of entries. Since $x_t$ carries little information beyond $m^*$:
\begin{align}
    \Delta I_{\text{compress}} &\approx 0 
    \quad \text{(no new entry added)}, \\
    \Delta I_{\text{retrieve}} &\approx 0 
    \quad \text{($m^*$ already covers $x_t$'s 
    contribution to $T$)}.
\end{align}
Merge is therefore a safe operation when redundancy is high: it does not degrade retrievability while keeping memory compact.
 
\paragraph{Append.}
When $x_t$ is largely independent of all entries in $M_t$, appending it as a new entry increases memory size but may capture evidence for future queries:
\begin{align}
    \Delta I_{\text{compress}} &= H(x_t | M_t) > 0 
    \quad \text{(memory grows)}, \\
    \Delta I_{\text{retrieve}} &= I(x_t; T | M_t, Q) 
    \quad \text{(new evidence for some query types)}.
\end{align}
Append is warranted when the retrievability gain outweighs the compression cost.
 
\paragraph{Link.}
When $x_t$ is semantically related to $m^*$ but carries distinct information, linking them preserves both entries while encoding their relationship through an associative edge. The edge metadata (relation type $\rho$ and connection strength $s$) provides an additional retrieval signal: during neighbor expansion, edge weights modulate score propagation, directing retrieval toward more informative paths. This increases $I(M, Q; T)$ for queries requiring multi-hop reasoning, where evidence is distributed across multiple entries.
 
\subsection{Candidate Selection}
\label{app:candidate_selection}
 
The gated construction (Section~\ref{sec:gated_update}) requires evaluating the semantic relation between $x_t$ and existing memory entries. Computing this for all entries is impractical, so \textsc{\method} uses a two-stage filtering approach:
 
\paragraph{Stage 1: Embedding-based candidate ranking.}
Candidate entries are ranked by cosine similarity between their context embeddings and $\mathbf{h}_t = \text{Embed}(c_t)$. This serves as a computationally efficient filter that identifies entries most likely to have high semantic overlap with $x_t$.
 
\paragraph{Stage 2: LLM-based semantic evaluation.}
Top-ranked candidates are evaluated by the LLM, which classifies the semantic relation (redundant, complementary, or novel) and estimates connection strength $s \in [0, 1]$. This stage handles cases where embedding similarity alone is insufficient to determine the appropriate operation, such as when two entries discuss the same topic from different perspectives. 

\section{Prompting Templates}
\label{app:prompts}
This appendix provides the complete prompting templates used in \textsc{\method}. All prompts are designed to elicit structured JSON outputs for reliable parsing.
 
\subsection{Memory Construction Prompts}
 
\subsubsection{Semantic Ingestion Prompt}

During the ingestion phase (Section~\ref{sec:ingestion}), raw conversational input $x_t$ is transformed into a structured note $n_t = (r_t, c_t, \mathbf{h}_t, \mathcal{K}_t)$. The following prompt instructs the LLM to extract keywords $\mathcal{K}_t$ and generate the denoised context $c_t$:
 
\begin{tcolorbox}[colback=gray!5, colframe=gray!75, 
title=Semantic Ingestion Prompt, fontupper=\scriptsize,
breakable
]
You are an expert Knowledge Graph Extractor. Your task is to analyze the [TARGET TURN] to extract structured metadata.
 
\textbf{Input Text:} \{content\}
 
\textbf{Guidelines:}
 
\textbf{1. Keywords (Entities):}
\begin{itemize}[leftmargin=*, nosep]
    \item \textbf{GOAL}: Extract 3-5 specific Noun Phrases explicitly present in the text.
    \item \textbf{FOCUS PRIORITIES}:
    \begin{enumerate}[nosep]
        \item Proper Nouns: People (e.g., ``Melanie''), Locations, Organizations.
        \item Concrete Objects: Physical items (e.g., ``painting'', ``plate'', ``contract'').
        \item Specific Topics: ``LGBTQ support group'', ``deadline''.
    \end{enumerate}
    \item \textbf{CRITICAL STOP LIST}: Ignore conversational meta-roles, abstract terms, and speaker names acting purely as subjects.
    \item \textbf{ZERO-SHOT RULE}: If the text is purely phatic (e.g., ``Wow'', ``That's cool''), return an empty list.
\end{itemize}
\textbf{2. Context (Factual Restatement):}
\begin{itemize}[leftmargin=*, nosep]
    \item \textbf{GOAL}: Rewrite the text into a self-contained factual statement.
    \item \textbf{CONSTRAINT 1 (Strict Fidelity)}: Only use information present in the Input Text.
    \item \textbf{CONSTRAINT 2 (Safe Resolution)}: Resolve ``I/my/we'' using the Speaker's name if it appears in the text. For external pronouns where the antecedent is missing, keep the pronoun or use a generic term. Do not guess.
    \item \textbf{CONSTRAINT 3 (No Meta-Language)}: Start directly with the subject. Avoid ``The speaker says...''.
\end{itemize}
\vspace{0.5em}
\textbf{Output Format (JSON):}
\begin{verbatim}
{"keywords": ["entity1", "entity2"], "context": "Melanie thinks the item is cool."}
\end{verbatim}
\end{tcolorbox}
 
The extracted keywords are matched against the existing keyword index $\mathcal{K}$ to establish symbolic anchors, while the context is encoded via the embedding model to obtain $\mathbf{h}_t = \text{Embed}(c_t)$. This dual extraction enables both symbolic and semantic access pathways during retrieval.
 
\subsubsection{Gated Construction Prompt}
\label{app:gating}
 
During gated construction (Section~\ref{sec:gated_update}), the LLM evaluates the semantic relation between a new note $n_t$ and each candidate note $n_i \in \mathcal{N}_{\text{cand}}$. The following prompt classifies the relation type and estimates connection strength $s(n_t, n_i) \in [0, 1]$, determining the appropriate construction operation (Merge, Link, or Append).
 
\begin{tcolorbox}[colback=gray!5, colframe=gray!75, title=Gated Construction Prompt, fontupper=\scriptsize,
breakable]
\textbf{Role:} You are a Knowledge Graph Updater. Your job is to evaluate the relationship between a NEW NODE and existing CANDIDATE NODES.
 
\vspace{0.5em}
\textbf{[NEW NODE]}\\
Content: ``\{content\}''\\
Context: ``\{context\}''\\
Keywords: \{keywords\}
 
\vspace{0.5em}
\textbf{[CANDIDATE NODES]}\\
\{candidates\_str\}
 
\vspace{0.5em}
\textbf{Instructions:}
 
Analyze each candidate and generate a JSON response following these rules:
 
\textbf{1. Analyze Relationship:}
\begin{itemize}[leftmargin=*, nosep]
    \item Determine \texttt{relation\_type}: `SUPPORTS', `CONFLICTS', or `RELATED\_TO'.
    \item Assign \texttt{connection\_strength} (0.0 -- 1.0), indicating the degree of semantic overlap or logical connection.
\end{itemize}
 
\textbf{2. Determine Operation (Based on Strength):}
\begin{itemize}[leftmargin=*, nosep]
    \item \textbf{CASE A: Strength $\geq$ 0.7 (High Redundancy)} $\rightarrow$ \textbf{MERGE}
    \begin{itemize}[nosep]
        \item Action: Integrate details from the New Node into the candidate's context.
        \item Template: ``[Original Context]. Specifically, [New Node Info]...''
    \end{itemize}
    
    \item \textbf{CASE B: Strength $\in$ [0.5, 0.7) (Complementary)} $\rightarrow$ \textbf{LINK}
    \begin{itemize}[nosep]
        \item Action: Establish associative edge; keep contexts separate.
        \item Template: ``[Original Context]. (Related: [New Node Keyword])''
    \end{itemize}
    
    \item \textbf{CASE C: Strength $<$ 0.5 (Distinct)} $\rightarrow$ \textbf{APPEND}
    \begin{itemize}[nosep]
        \item Action: Add New Node as autonomous unit; no modification.
    \end{itemize}
    
    \item \textbf{CASE D: relation\_type is `CONFLICTS'} $\rightarrow$ \textbf{LINK with Contrast}
    \begin{itemize}[nosep]
        \item Action: Note the conflicting information explicitly.
        \item Template: ``[Original Context]. However, [New Node] indicates that...''
    \end{itemize}
\end{itemize}
 
\textbf{3. Output}: Return strictly valid JSON matching the schema.
\end{tcolorbox}
 
\vspace{0.5em}
\noindent\textbf{Mapping to Paper Notation.} The \texttt{connection\_strength} output corresponds to $s(n_t, n_i) \in [0, 1]$ in Section~\ref{sec:gated_update}, and \texttt{relation\_type} corresponds to $\rho \in \{\textsc{Supports}, \textsc{Conflicts}, \textsc{Related\_To}\}$. The thresholds specified in the prompt (0.7 for merge, 0.5 for link) guide the LLM's classification into the three operation regimes defined in Eq.~\eqref{eq:update_rule}. When the Merge operation is triggered, the LLM generates a unified context $c'_i = \mathcal{F}_{\text{merge}}(c_i, c_t)$ following the template specified in Case A, preserving all distinct information from both notes while reducing redundancy. Full hyperparameter settings are provided in Appendix~\ref{app:hyperparameters}.

\subsection{Memory Retrieval Prompts}
 
\subsubsection{Query Intent Analysis Prompt}
\label{app:query_intent}
 
During the query decomposition step (Section~\ref{sec:retrieval}), the raw query $q$ is processed by a semantic parser to extract retrieval signals. The following prompt extracts a topical description and entity keywords for driving the topic and keyword channels.
 
\begin{tcolorbox}[colback=gray!5, colframe=gray!75, title=Query Intent Analysis Prompt, fontupper=\scriptsize,
breakable]
Analyze the user query to identify its \textbf{Target Taxonomy Category}, extract key entities, and detect time-related intent.
 
\vspace{0.5em}
\textbf{Task 1: topic\_desc (Target Category)}
\begin{itemize}[leftmargin=*, nosep]
    \item Predict the \textbf{Taxonomy Category} or \textbf{Subject Heading} this query falls under.
    \item \textbf{Style}: Strict Noun Phrase (like a book chapter title or library category).
    \item \textbf{Constraint}: Keep it under 8 words.
    \item Do NOT describe the user's intent (e.g., avoid ``how to...'', ``techniques for...''). Instead, name the topic itself.
\end{itemize}
 
\textbf{Task 2: Keywords}
\begin{itemize}[leftmargin=*, nosep]
    \item Extract 3-5 core entities, technical terms, or specific concepts.
    \item \textbf{CRITICAL}: Convert terms to their canonical singular form (e.g., ``transformers'' $\rightarrow$ ``Transformer'').
    \item Exclude generic verbs or stop words.
\end{itemize}
\vspace{0.5em}
\textbf{Query:} \{query\}
 
\vspace{0.5em}
\textbf{Output Format (JSON):}
\begin{verbatim}
{"topic_desc": "Concise Noun Phrase",
 "keywords": ["keyword1", "keyword2", "keyword3"]}
\end{verbatim}
\end{tcolorbox}
 
\vspace{0.5em}
\noindent\textbf{Mapping to Paper Notation.} The \texttt{topic\_desc} field is encoded via the embedding model to obtain $\mathbf{h}_{\text{topic}} \in \mathbb{R}^d$, which drives the topic channel described in Section~\ref{sec:retrieval}. The \texttt{keywords} are similarly embedded to form $\mathbf{H}_{\text{keys}} = \{\mathbf{h}_{k_1}, \ldots, \mathbf{h}_{k_m}\}$, driving the keyword channel.
 
\subsubsection{Iterative Evidence Refinement Prompts}
\label{app:ier}
 
The Iterative Evidence Refinement protocol (Section~\ref{sec:iterative_refinement}) employs two complementary prompts: a \textit{Sufficiency Evaluator} that assesses whether the current evidence pool adequately addresses the query, and a \textit{Sub-query Generator} that synthesizes targeted follow-up queries when gaps are identified.
 
\vspace{0.5em}
\noindent{\textbf{Sufficiency Evaluation Prompt.}} At each iteration $i$, the following prompt evaluates whether the current evidence pool $\mathcal{E}^{(i)}$ sufficiently addresses the query.
 
\begin{tcolorbox}[colback=gray!5, colframe=gray!75, title=Sufficiency Evaluation Prompt, fontupper=\scriptsize,
breakable]
You are a reflector agent that evaluates whether the current context and answer is sufficient to answer a question.
 
\vspace{0.5em}
\textbf{Question:} \{question\}
 
\vspace{0.5em}
\textbf{Current Context and Current Answer:}\\
\{context\}
 
\vspace{0.5em}
Evaluate whether the provided context and answer contains enough information to answer the question comprehensively.
 
If the context and answer is insufficient, identify what specific information is missing.
 
\vspace{0.5em}
\textbf{Output Format (JSON):}
\begin{verbatim}
{"sufficient": true or false,
 "missing_info": "description of missing information",
 "confidence": 0.0 to 1.0}
\end{verbatim}
\end{tcolorbox}
 
\vspace{0.5em}
\noindent\textbf{Sub-query Generation Prompt.} When the sufficiency evaluation returns \texttt{sufficient: false}, the following prompt generates a refined sub-query $q^{(i+1)}$ targeting the identified information gaps.
 
\begin{tcolorbox}[colback=gray!5, colframe=gray!75, title=Sub-query Generation Prompt, fontupper=\scriptsize,
breakable]
You are a Query Evolution Agent. Your goal is to decompose a complex user question into a specific, actionable sub-query to retrieve missing information from a Knowledge Graph.
 
\vspace{0.5em}
\textbf{Original Question:} ``\{query\_str\}''
 
\vspace{0.5em}
\textbf{Current Known Information (Context):}\\
\{context\_str\}
 
\vspace{0.5em}
\textbf{History of Reasoning Steps (Q\&A):}\\
\{prev\_reasoning\}
 
\vspace{0.5em}
\textbf{CRITICAL}: The Reflector Agent has identified the following MISSING INFORMATION needed to answer the main question:\\
\{missing\_info\}
 
\vspace{0.5em}
\textbf{Task:} Based on the ``Missing Information'' and ``History'', formulate the \textbf{NEXT} single sub-question to retrieve this missing info.
\begin{itemize}[leftmargin=*, nosep]
    \item The sub-question must be specific.
    \item It should act as a search query for the next hop.
    \item If we have enough information to answer the main question, return `None'.
\end{itemize}
 
\vspace{0.5em}
\textbf{Sub-question:}
\end{tcolorbox}
 
\vspace{0.5em}
\noindent\textbf{IER Protocol Flow.} The two prompts work in tandem: the Sufficiency Evaluator determines whether to terminate (when \texttt{sufficient: true} or iteration count reaches $I_{\max}$), while the Sub-query Generator drives evidence expansion by producing targeted queries that are re-executed through the tri-pathway retrieval mechanism (Section~\ref{sec:retrieval}). The \texttt{confidence} score from the Sufficiency Evaluator can optionally be used for early termination when confidence exceeds a predefined threshold.

\section{Dataset Statistics}
\label{app:dataset}
Table~\ref{tab:locomo} presents the sample distribution across the five reasoning categories. The categories are designed to test distinct memory capabilities: \textit{Multi-Hop} requires synthesizing evidence across multiple memory units; \textit{Temporal} tests reasoning about time-dependent information and event ordering; \textit{Open Domain} evaluates retrieval of general knowledge from conversation history; \textit{Single Hop} assesses precise entity matching and direct fact retrieval; and \textit{Adversarial} challenges the system with distractors and misleading information.

As shown in Table~\ref{tab:locomo}, the category distribution is imbalanced, with Single Hop comprising the largest proportion (42.3\%) and Open Domain the smallest (4.8\%). To account for this imbalance, the average scores reported in Table~\ref{tab:main_gpt} and Table~\ref{tab:main_qwen} are computed as \textbf{weighted averages} based on category sample sizes, ensuring that performance on larger categories contributes proportionally to the overall evaluation.


\begin{table}[h]
\centering
\caption{LoCoMo benchmark category distribution.}
\label{tab:locomo}
\begin{tabular}{lcc}
\toprule
Category & Samples & Proportion \\
\midrule
Multi-Hop & 282 & 14.2\% \\
Temporal & 321 & 16.2\% \\
Open Domain & 96 & 4.8\% \\
Single Hop & 841 & 42.3\% \\
Adversarial & 446 & 22.5\% \\
\midrule
Total & 1,986 & 100\% \\
\bottomrule
\end{tabular}
\end{table}

\section{Hyperparameter Settings}
\label{app:hyperparameters}
Table~\ref{tab:hyperparams} summarizes all hyperparameters used in \textsc{\method}. These hyperparameters are organized by the two main phases of our framework: memory construction and memory retrieval.

For memory construction, the merge threshold $\tau_m = 0.7$ and link threshold $\tau_l = 0.5$ control the gated structural update decisions (Eq.~\eqref{eq:update_rule}). A higher merge threshold ensures that only highly redundant information is consolidated, preserving fine-grained distinctions between memory units. The link threshold is set lower to capture complementary relationships that support multi-hop reasoning.

For memory retrieval, we set $K_{topic} = 3$ to balance navigation precision with coverage, allowing the system to explore multiple relevant Topic clusters. The keyword retrieval parameter $K_{key} = 10$ provides sufficient anchor points for entity-centric queries. The final pool size $K_{final} = 20$ bounds the evidence passed to the generation stage, balancing context richness against computational cost. The maximum IER iterations $I_{max} = 3$ prevents excessive retrieval loops while allowing sufficient evidence expansion for complex queries.

All hyperparameters were tuned on a held-out validation set. We found the framework to be relatively robust to moderate variations in these values, with performance degrading gracefully when parameters deviate within $\pm 20\%$ of the reported settings.

\begin{table}[htbp]
\centering
\caption{Hyperparameter settings.}
\label{tab:hyperparams}
\begin{tabular}{llc}
\toprule
Phase & Parameter & Value \\
\midrule
\multirow{2}{*}{Construction} 
& Merge threshold $\tau_m$ & 0.7 \\
& Link threshold $\tau_l$ & 0.5 \\
\midrule
\multirow{4}{*}{Retrieval}
& Topic retrieval $K_{topic}$ & 3 \\
& Keyword retrieval $K_{key}$ & 10 \\
& Final pool size $K_{final}$ & 20 \\
& Max IER iterations $I_{max}$ & 3 \\
\midrule
\multirow{2}{*}{Generation}
& Temperature (general) & 0.7 \\
& Temperature (adversarial) & 0.5 \\
\bottomrule
\end{tabular}
\end{table}

\section{Compute Resources}
All experiments were conducted on a single server equipped with Intel Xeon Platinum 8480C CPUs and NVIDIA H800 80GB GPUs. Memory construction, retrieval, and generation for all backbone models and baselines were executed on this hardware. The primary computational cost arises from LLM-based semantic evaluation during memory construction (Section \ref{sec:online_consolidation}), which requires one LLM call per candidate pair for gated classification. For closed-source models (GPT-4o, GPT-4o-mini), construction and generation costs are dominated by API latency rather than local compute.

\section{Broader Impact}
This work improves memory systems for LLM-based agents, enabling more effective evidence retrieval from long-term interaction histories, which can benefit applications such as personal assistants and educational tools. However, enhanced long-term memory raises privacy concerns, as systems that consolidate detailed user interaction records may inadvertently expose sensitive information if not properly governed. The associative linking mechanism may also surface unintended connections between pieces of personal information. Additionally, the LLM-based construction components inherit biases from the underlying models, which could propagate through the memory hierarchy to downstream reasoning. We encourage practitioners to implement data governance policies, user-controlled forgetting mechanisms, and bias auditing when deploying such systems in practice.



\end{document}